\documentclass[runningheads,a4paper]{llncs}

\usepackage{amssymb}
\usepackage{graphicx}
\usepackage{cite}
\usepackage{url}
\usepackage{comment}
\urldef{\mailsa}\path|benrjabamal.ihec@gmail.com|
\urldef{\mailsb}\path| Mouloud.Kharoune@univ-rennes1.fr|
\urldef{\mailsc}\path|zoltan.miklos@irisa.fr|
\urldef{\mailsd}\path|Arnaud.Martin@univ-rennes1.fr|
\newcommand{\keywords}[1]{\par\addvspace\baselineskip
\noindent\keywordname\enspace\ignorespaces#1}
\usepackage[cmex10]{amsmath}
\usepackage[linesnumbered,ruled,vlined]{algorithm2e}
\begin{document}

\mainmatter  

\title{Characterization of experts in crowdsourcing platforms} 

\titlerunning{Characterization of experts in crowdsourcing platforms}

\author{Amal Ben Rjab${}^1$ \and Mouloud Kharoune${}^2$ \and Zoltan Miklos${}^2$ \and  Arnaud Martin${}^2$ }

\authorrunning{Characterization of experts in crowdsourcing platforms}

\institute{
        ${}^1$        \small{University of Tunis/LARODEC laboratory, Tunisia}  \\
        ${}^2$        \small{University of Rennes 1/ IRISA, France }\\
\mailsa\\
\mailsb\\
\mailsc\\
\mailsd\\}
%

\toctitle{Lecture Notes in Computer Science}
\tocauthor{Authors' Instructions}
\maketitle

\begin{abstract}
\emph{
Crowdsourcing platforms enable to propose simple human intelligence tasks to a large number of participants who realise these tasks. The workers often receive a small amount of money or the platforms include some other incentive mechanisms, for example they can increase the workers reputation score, if they complete the tasks correctly. We address the problem of identifying experts among participants, that is, workers, who tend to answer the questions correctly. Knowing who are the reliable workers could improve the quality of knowledge one can extract from responses.  As opposed to other works in the literature, we assume that participants can give partial or incomplete responses, in case they are not sure that their answers are correct. We model such partial or incomplete responses with the help of belief functions, and we derive a measure that characterizes  the expertise level of each participant. This measure is based on  precise and exactitude degrees that represent two parts of the expertise level. The precision degree reflects the reliability level of the participants and the exactitude degree reflects the  knowledge level of the participants. We also analyze our model through simulation and demonstrate that our richer model can lead to more reliable identification of experts.}

\end{abstract}

\section{Introduction} \label{sec: introduction}
Crowdsourcing is term for  {\em ``the act of a company or institution taking a function once performed by employees and outsourcing it to an undefined (and generally large) network of people in the form of an open call"}~\cite{howe2006rise}. Crowdsourcing platforms are used more and more often to execute tasks that are hard for computers but easy for humans.  This form of realizing small human intelligence tasks through a large number of individuals has been used in various domains; and plays a more and more important role.  It  is also  considered as a style of future work~\cite{kittur2013future} that can be crucial for example  in the context of decision support~\cite{chiu2014can}. 
Controlling the quality of obtained data and identifying the workers  who tend to give correct answers in this environment still a major problem. The absence of quality control of participants (and their responses) reduces the efficiency of these platforms~\cite{ipeirotis2010quality}.

One often refers to a participant who gives exact and precise answers as an expert~\cite{rjabcaracterisation}. Several works~\cite{ipeirotis2010quality}~\cite{khattak2011quality,kazai2011worker,Noll2009} were proposed to identify the experts in this context. These methods assume that if a worker accepts to complete a task, he will give an answer, even if he is not sure about it. In other words, they make the assumption that a worker does not skip a question. Also, existing crowdsourcing platforms do not allow to give partial results.  For example, if the tasks involve a multiple choice question with answers $A$, $B$, $C$ and $D$, a worker cannot say that the correct answer either $A\, or\, B$ (he is not sure about), but certainly not $C\, or\, D$.   

Some works use first ``gold"  data on which real answers are known~\cite{google}. In that case, a degree of exactitude (the percentage of answers that is not wrong) and a degree of precision (the percentage of answers that is not partial) could be learn to measure the expertise level. Here, we assume we that  do not have such data.

In our work, we construct a model where we allow situations where a worker skips some questions or answers them partially. In our model we make use of belief functions that is a powerful framework to take into account such imperfection of data. We propose a novel expert identification technique that by calculating a degree of exactitude (based on a level of answers that is not wrong) and a degree of precision (based on a level of answers that is not partial). The ``ideal'' worker has a high degree of exactitude and a high degree of precision. For example, in the multiple choice question case, if the correct answer is $A$ then clearly the answer $A$ is better than an answer $A \, or \, B$ (higher degree of precision).

The degrees of exactitude and precision are complementary, so using both of them together can lead to better expert identification methods. The rest of the paper is organized as follows.  Section~\ref{sec: crowdsourcing} formulates the expert identification problem more precisely, together with some relevant related work.  We present our approach in Section~\ref{sec: expert identification}. The experimental evaluation is presented in Section~\ref{sec:experimentation}.

\section{Expert identification in the context of crowdsourcing}\label{sec: crowdsourcing}

\subsection{Notions of an expert}
An expert in the context of crowdsourcing, is the person who provides  a large number of correct, complete and reliable answers. 
The  person who acquired a set of knowledge and skills about a particular area.  He can extract  knowledge and  relevant responses  with a minimum cognitive effort.
He is identified in crowdsourcing platforms  by:  
the precision and the exactitude of responses, the capability to detect the tasks {\em a priori}, the knowledge, skills  and learning level.

\subsection{Expert identification methods} 
Evaluating quality of workers and identifying experts in  crowdsourcing represents a standing problem. Many authors found that taking randomly  workers is a good choice~\cite{bozzon2013choosing} and others found that establishing a good strategy for selecting experts is more interesting~\cite{ipeirotis2010quality}. Several researches have been exploring this area, but essentially there are two basic approaches to identify the experts: 
\textbf{\textit {Use ``gold" data:}} Provide participants the questions that we already know the answers and  identify the  workers who give the correct responses as the experts. 
\textbf{\textit {Use multiple workers:}} Give a score for each participant which represents his qualities and skills.
In this context, Ipeirotis {\em et al.}  improved in~\cite{ipeirotis2010quality} the expectation maximization algorithm (EM) 
 to generate a scalar score representing the quality of each worker.  \cite{khattak2011quality} proposed an evaluation of the participants by the set of labels. 
\cite{kazai2011worker} based on behavioral observation to define a typology of workers.   

\cite{Noll2009}  proposed an algorithm based on the graphs (SPEAR)
  to classify the users and to identify the experts.
Various methods proposed to identify the experts. But,  all these methods have a  such level of imprecision and inaccuracy results. In order to ensure a certain identification,  we propose to model this imperfection. We proposed an identification of  experts with using the theory of  belief functions~\cite{dempster1967upper,shafer1976mathematical} which represents a mathematical theory   for representing imperfect information and gives a complete framework to model the participant's answers.

\section{Identification of the experts}\label{sec: expert identification}

We would like to identify the experts in a crowdsourcing platform. We assume that the questions (tasks) and a list of answers from the crowd workers available. However, we do not assume any access to a ``gold'' data that would contain all the correct answers. Such a ground truth would clearly largely simplify the identification of experts. Therefore, we develop novel techniques -based on the theory of belief functions- to calculate the exactitude and precision degrees.  
   
We use the following formalism. We note the responses $r_{U_{j}}$ proposed by each participant $U_{j}$ with a mass of belief $m_{U_{j}}^{\Omega_{k}}$. Each response is specific for each question $Q_{k}$ ($k=\{1,\cdots,K\}$) which has a specific frame of discernment $\Omega_{k}$ with $\Omega_{k}=\{\omega_{1}^{Q_{k}},\ldots,\omega_{n_k}^{Q_{k}}\}$. The frame $\Omega_k$ is the set of all possible responses of $Q_k$ question.  Therefore, we obtain a matrix of mass of belief of size $s$ participants/lines and $K$ questions/columns given by:
\begin{eqnarray}
\label{matrice_info}
\left.
\begin{array}{cc} 
& \left.
\begin{array}{ccccc}
Q_1& \ldots & Q_k &  \ldots & Q_K \\
\end{array}
\right. \\
\left.
 \begin{array}{c}
  U_1 \\ \vdots \\ 
  U_j \\ \vdots \\ U_s\\
\end{array}
\right.
& \left[
\begin{array}{ccccc}
\displaystyle{m_{U_{1}}^{\Omega_{1}}} & \ldots & \displaystyle{m_{U_{1}}^{\Omega_{k}}}& \ldots & \displaystyle{m_{U_{1}}^{\Omega_{K}}} \\
\vdots & &\vdots & & \vdots \\ 
 \displaystyle{m_{U_{j}}^{\Omega_{1}}} & \ldots & \displaystyle{m_{U_{j}}^{\Omega_{k}}}& \ldots & \displaystyle{m_{U_{j}}^{\Omega_{K}}} \\
\vdots & & \vdots & & \vdots \\
 \displaystyle{m_{U_{s}}^{\Omega_{1}}} & \ldots & \displaystyle{m_{U_{s}}^{\Omega_{k}}} & \ldots & \displaystyle{m_{U_{s}}^{\Omega_{K}}} \\
\end{array}
\right]
\end{array}
\right.
 \end{eqnarray}

\subsection{Exactitude degree}
The exactitude degree is based on the average of the distance between the response proposed by the participant $m_{U_{j}}^{\Omega_{k}}$ and all the responses of the other participants $m_{U_{\varepsilon_{s-1}}}^{\Omega_{k}}$. This representation of all other participants is obtained by the average of the responses proposed by the $s-1$ participants for the $k^{th}$ question, such as:
\begin{eqnarray}
m_{U_{\varepsilon_{s-1}}}^{\Omega_{k}}(X)=
	\displaystyle \frac{1}{s-1}\!\! \sum_{j=1}^{s-1}  m_j(X)
\end{eqnarray}
The distance is then calculated by the distance of Jousselme~\cite{jousselme2001new}: $d_J(m_{U_{i}}^{\Omega_{k}}, m_{U_{\varepsilon_{s-1}}}^{\Omega_{k}})$. 
According to this distance, we  calculate the exactitude degree for each participant $U_{j}$ as  follows:
\begin{equation}
IE_{U_{j}}=1-\frac{1}{r_{(U_{j})}}\displaystyle{\sum_{k=1}^{K}d_{U_{j}}^{\Omega_{k}}}
\end{equation}
The assumption behind this method is the majority of participants give a correct answer. This assumption is currently made in information fusion and crowdsourcing.

The exactitude degree can be used to identify the experts. For this purpose, we use the $k$-means algorithm (with $k=2$ for expert/non expert). The set of experts is given by the cluster with the higher average of exactitude degree.

\subsection{Precision degree}
Based on the model of responses given by the mass functions $m_{U_{j}}^{\Omega_{k}}$, we can define a degree of precision. 

We recall that we allow the participants to give partial answers, that is crucial for calculating the precision degree. The usual model of responses (that is, the worker must give a complete answer), we could not  define a such degree. 

We note $\displaystyle{\delta_{U_{j}}^{\Omega_{k}}}$ the specificity degree of the mass function $m_{U_{j}}^{\Omega_{k}}$.  It  is defined by~\cite{Smarandache11a} as follows:
\begin{equation}
\delta_{U_{j}}^{\Omega_{k}} =1- \displaystyle{\sum_{X\in2^{\Omega_{k}}} m_{U_{j}}^{\Omega_{k}}(X){\frac{ \log_2 (|X|)}{\log_2 (|\Omega_{k}|)}}}
\end{equation}
This specificity degree allows to translate the precision  level of each response independently of the other participant's responses. 
To measure the degree of precision of each participant $IP_{U_{j}}$, we propose to calculate the average of the specificity degrees for all the $k^{th}$ questions.  Such as:
\begin{equation}
IP_{U_{j}}=\frac{1}{r_{(U_{j})}}\displaystyle{\sum_{k=1}^{K}\delta_{U_{j}}^{\Omega_{k}}}
\end{equation}
We determine the experts by using  $k$-means (with $k=2$). We do not need the assumption on the majority of participant's answers.

\subsection{Global degree}
In order to obtain a global degree, we combine both degrees in a single degree for each participant. The global degree is given by a weighted average as follows:
\begin{equation}
\label{global_degree}
GD_{U_{j}}=\beta_{U_{j}} IE_{U_{j}} +(1- \beta_{U_{j}}) IP_{U_{j}}
\end{equation}
The weight $\beta_{U_{j}}$ is introduced to give more or less importance for each degree. Hereafter, we do not make any difference between the participants in the crowd.


\section{Experimentation} \label{sec:experimentation}

In the following, we generate some mass functions in order to evaluate our approach in the context where there is not use of gold data.
We generate three kinds of participants. The {\bf experts} are those who provide precise and exact responses, in the generation of the masses a {\em singleton} is expected on the correct answer. However, if the expert is not totally sure of him, the {\em ignorance} is also a focal element. The {\bf imprecise experts} are those who provide exact but imprecise answers, the correct {\em singleton} can be in a disjunction and the {\em ignorance} can also be a focal element. The {\bf ignorants} (sometimes called spammers) are those who give random responses with mass functions taken randomly.  
To verify the efficiency of our approach we make several experiments with 100 participants, 100 questions where each experiment is repeated 10 times.

\begin{figure}
\centering
\includegraphics[width=9cm,height=65mm]{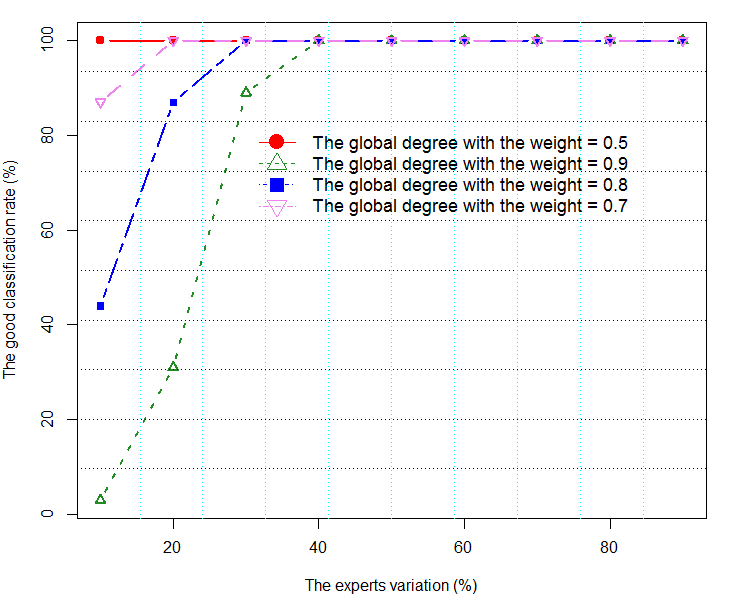}
\caption{Variation of the good classification rate according to the percentages of experts}
\label{fig:variation4}
\end{figure}

The precision or the exactitude degree alone is insufficient to identify the experts. The global degree of the equation~\eqref{global_degree} allows to identify precise and exact responses simultaneously. In a first  experiment (with results illustrated in Figure~\ref{fig:variation4}), we vary the experts' number, without generating imprecise experts, from 10~\% to 90~\% with the global degree in order to prove the ability of our method to identify precise and exact responses simultaneously. 
In order to demonstrate the importance of each degree we vary in each case the weight $\beta_{U_{j}}$ from 0.1 to 0.9. 100~\% Good classification rate with $\beta_{U_{j}} = 0.5$ reflects that both exactitude and precision degrees have the importance to identify experts. Our algorithm identifies correctly the experts and puts all the other participants in the class of the ignorant. 

\begin{figure}[!h]
\centering
\includegraphics[width=9cm,height=60mm]{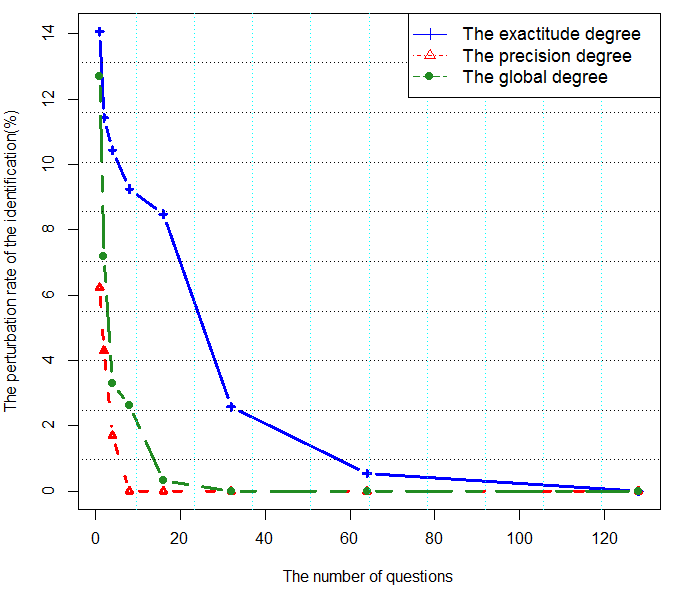}
\caption{The variation of the perturbation rate according to the different degrees}
\label{fig:variation5}
\end{figure}

To verify the stability of the good classification rates, we vary in the next experiment (with results illustrated in Figure~\ref{fig:variation5}) the number of questions with 35~\% of experts, 35~\% of imprecise experts and 30~\% of ignorants for 10 iterations, we calculate the three degrees. We measure this stability with a perturbation rate calculated by the standard deviation between the different good classification rate exchange on 10 iterations.
This experiment shows that it is necessary to have a certain number of questions in order to ensure a better identification. 

\begin{figure}[!h]
\centering
\includegraphics[width=9cm,height=60mm]{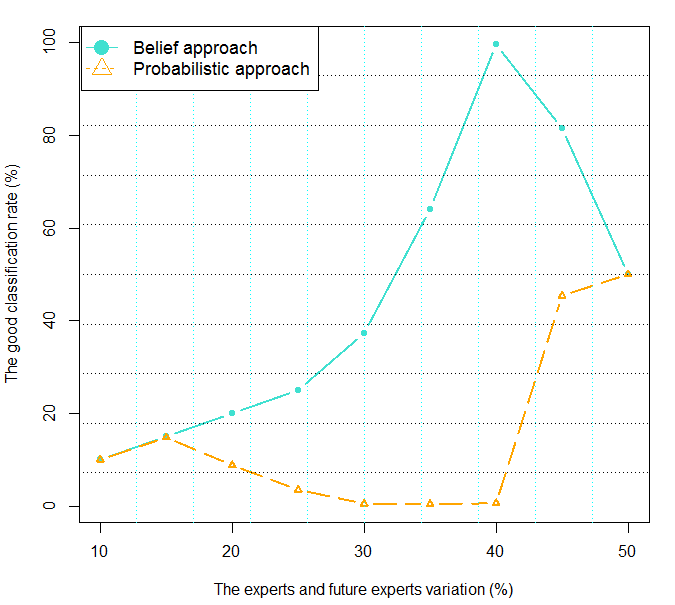}
\caption{Comparison between belief function and  probability function}
\label{fig:variation6}
\end{figure}

We can found that 30 questions provide a reliable good classification rate. All the previous experiments show the ability of our method to identify the experts in the context of uncertain and imprecise responses. The recourse to the theory of belief functions ensures a reliable identification. It solves the problem of imperfection and provides a certain frame of characterization. With both degrees, we detect the exactitude and precision level of each participant and we correctly identify the experts in the crowd. 
To confirm the interest of the theory of the belief functions, we compare our belief approach with the probabilistic approach corresponding to the mass function $m_{U_j}^{\Omega_k}$ which models the responses proposed by each participant $U_{j}$ given by the pignistic probability:
\begin{equation}
 BetP_{m_{U_j}^{\Omega_k}}(\omega_{k})= \displaystyle \sum_{X \subseteq \Omega_k , \omega_k \in X} \frac{m_{U_j}^{\Omega_k}(X)}{(1-m_{U_j}^{\Omega_k}(\emptyset)) |X|}
\end{equation}
With the same principle in section~\ref{sec: expert identification}, we calculate the exactitude degree as follows:
\begin{equation}
EP(U_{j})=1-\frac{1}{r_{(U_{j})}}\displaystyle{\sum_{k=1}^{K}d_{U_{j}}^{\Omega_{k}}}
\end{equation}
Where $d_{U_{j}}^{\Omega_{k}}$ is the Euclidean distance on the probabilities. We have to do the same assumption on the majority of correct answers. We use $k$-means to characterize the experts.  In this way, we obtain a probabilistic approach available to detect experts.  
We limit the comparison by the exactitude degree, due to the impossibility to determine the specificity degree with the probability. 
We vary in this experiment the percentage of experts and imprecise experts at the same time. 
The results are illustrated in Figure~\ref{fig:variation6}.
This figure shows the interest of the use of the belief functions theory to identify the experts and imprecise experts.  The probabilistic approach cannot identify  the experts from the  imprecise experts, it  loose the information of exactitude and could not model the imprecision. 
The regression of the good classification rate to 0~\%  reflects this  inability. 
Whereas with the  belief  approach the precise and imprecise experts are better discriminated with all the variations.  In  complex environment like the crowdsourcing, the theory of belief functions can consider all the imperfection of the participant's responses.


\section{Conclusion}\label{sec: conclusion}
We introduced a new technique for characterizing the experts in a crowdsourcing platform by using the belief functions theory, to improve the quality of data that one could obtain from such platforms. We use a model where the crowd workers are allowed to skip a question or provide partial answers. Based on a belief model of the participant's responses, we calculated two complementary degrees: An exactitude degree translates the  knowledge level of the participants and a  precision degree  reflects their reliability  level.  We showed the ability of these degrees to help for the expert identification and we demonstrated the interest of the theory of  the belief functions in a comparison with the probability theory.

\bibliographystyle{IEEEtran}

\bibliography{my_bib_Belief}

\end{document}